\documentclass[11pt]{article}
\usepackage{coling2020}
\usepackage{times}
\usepackage{url}
\usepackage{latexsym}

\usepackage{graphics}
\usepackage{graphicx}
\usepackage{xcolor}
\usepackage{hyperref}
\usepackage{listings}
\usepackage[T1]{fontenc}
\usepackage[scaled=0.85]{beramono}

\colingfinalcopy 

\title{Provenance for Linguistic Corpora Through Nanopublications}

\author{Timo Lek$^1$, Anna de Groot$^1$, Tobias Kuhn$^1$, Roser Morante$^2$\\
$^1$Department of Computer Science,  Faculty of Science, Business Web and Media\\
$^2$CLTL Lab, Faculty of Humanities\\
Network Institute,  VU Amsterdam \\
 {\tt \{timolek97,annagrotius\}@gmail.com, \{t.kuhn,r.morantevallejo\}@vu.nl} 
 }

\date{}

\begin{document}
\maketitle

\newcommand{\todo}[1]{\noindent{\normalfont\color{red}\textbf{[}#1\textbf{]}}}

\maketitle              
\begin{abstract}
  Research in Computational Linguistics is dependent on text corpora for training and testing new tools and methodologies. While there exists a plethora of annotated linguistic information, these corpora are often not interoperable without significant manual work. Moreover, these annotations might have evolved into different versions, making it challenging for researchers to know the data's provenance. 
  This paper addresses this issue with a case study on event annotated corpora and by creating a new, more interoperable representation of this data in the form of nanopublications. We demonstrate how linguistic annotations from separate corpora can be reliably linked from the start, and thereby be accessed and queried as if they were a single dataset. We describe how such nanopublications can be created and demonstrate how SPARQL queries can be performed to extract interesting content from the new representations. 
  The queries show that information of multiple corpora can  be retrieved more easily and effectively because the information of different corpora is represented in a uniform data format. 

\end{abstract}

\section{Introduction}


The availability of annotated linguistic corpora is crucial to develop Natural Language Processing (NLP) systems \cite{chiarcos2012interoperability,van2018resource}. 
In the past couple of decades, many linguistic  resources have been released, which contain different types of annotations, from morphosyntactic to semantic and discourse level annotations. However, they often lack structural and conceptual interoperability \cite{chiarcos2012interoperability}. The former is related to the data format of annotations, such as XML-standoff or RDF representations, while the latter refers to the possibility of exchanging  information in a consistent and mappable way. Diverging annotation guidelines, the structure of annotated output, and the different amounts of annotations created per corpus pose a challenge in organizing the provenance of corpora. 

Often,  annotation projects run on (subsets of) existing text corpora, which are thereby annotated with multiple annotation layers. Unfortunately, the format for linking  annotations to the texts is mostly not standardized and these links therefore are often not fully precise or directly interoperable. This makes it challenging for a researcher to figure out the data's exact provenance and to integrate annotations from several corpora. Integrating different annotations is valuable because it can help scholars study and understand the different interdependencies between annotation layers (e.g. semantic parsers usually take syntactic structure in consideration). 




Moreover, often the annotations follow idiosyncratic guidelines that pursue a specific annotation goal.  The need to organize the various existing annotations has been recognized by the corpus linguistic community \cite{chiarcos2012interoperability,kilgarriff2001comparing}.
While a number of solutions to improve linguistic Linked Data have been presented in the past, we propose here 
a novel approach that uses the Linked Data format of nanopublications to represent annotated corpora in a more fine-grained semantic format, with the aim of boosting the interoperability of linguistic corpora and facilitating the automatic integration of annotations in a reliable and transparent manner. 

Our research is motivated by the empirical analysis on 20 event-annotated corpora presented in  \cite{van2018resource}, which highlighted the challenges, as well as the opportunity, of event corpora interoperability. Existing event annotation standards range from annotations about the participants, timing/temporal order, factuality, and entity coreference relation, making a standardized description of an event difficult. This is where nanopublications as a fine-grained and provenance-aware data format could provide transparency and interoperability for computational linguistic research.

\begin{figure}[htb]
    \centering
    \includegraphics[width=0.8\textwidth]{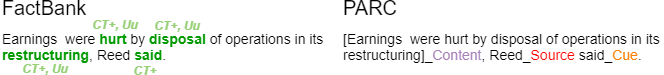}
    \caption{An example sentence with its annotations in FactBank and PARC.
    }
    \label{annotation_ex}
\end{figure}

For our case study, we have chosen to focus on integrating annotations from two corpora, FactBank \cite{sauri2009FactBank} and PARC 3.0 \cite{pareti2016parc}.  In PARC, attribution relations and their components (Cue, Content, Source) are annotated, while in FactBank events and their factuality values (i.e. the certainty of the occurrence of an event) are annotated. It is interesting to combine the two annotation layers in order to extract, for example, which events occur in the content of attribution relations and what are their factuality values. Figure \ref{annotation_ex} presents the same sentence annotated in PARC and FactBank.

Our  main contributions thereby are
(1) providing a fine-grained and interoperable Linked Data representation of annotated linguistic corpora in the form of nanopublications,
(2) analysing the main interoperability problems that we encountered and solved with our nanopublication-based approach, and
(3) demonstrating how the concrete annotations from our case study of two corpora can be integrated with nanopublications and queried with SPARQL.

We describe below how we converted these corpora into nanopublications and then report on the interoperability challenges faced and solved with our approach by representing the data in a uniform manner using nanopublications. Finally, we discuss based on our results the extent to which nanopublications can be seen as a viable alternative for improved linguistic corpora interoperability.

\section{Linked Data, Data Provenance and Nanopublications}

Linked Data is about connecting and publishing data in a structured way on the web, often on the basis of an ontology using the Web Ontology Language (OWL) \cite{mcguinness2004owl}. The data is often structured in triples using the Resource Description Framework (RDF). By using unique URIs for each entity of information, these entities can be interlinked in order to create an entire web of data \cite{bizer2011linked}.

Several projects were already performed to create linked linguistic data. Since 2012, six series of Linked Data in Linguistics (LDL) workshops have been hosted to gather and discuss contributions promoting linking linguistic data. After the second workshop, one of the presented efforts was the Linked Linguistic Open Data (LLOD) cloud.\footnote{\url{https://linguistic-lod.org/}}While creating the LLOD cloud, a lot of linguistic data was already available, which were mostly represented using RDF. One example is WordNet, which is also part of the Semantic Web \cite{gangemi2003ontowordnet,chiarcos2013linguistic}. Furthermore, Chiarcos proposed to represent linguistic corpora into OWL and RDF to interlink all of the different resources of corpora. In this way, the annotations of the corpora could be linked using terminological resources like standardized annotation formats \cite{chiarcos2012interoperability}.

Existing approaches that aim to represent the structure and content of natural language and its annotations include the Ontology Lexicalisation and the Ontologies of Linguistic Annotation (OLiA)
\cite{villegas2015parole,buitelaar2011ontology,bosque2019lemon}.
OLiA focuses on annotations of linguistic corpora and can be seen as a source that references to the `annotation terminology' that can be used in combination with the LLOD to capture all of the information of the annotations in the LLOD in a similar format \cite{chiarcos2015olia}.

Since all different texts and annotations in a corpus may have a different origin, i.e. provenance, provenance ontology PROV\footnote{\url{https://www.w3.org/ns/prov}} can be used for this. This ontology contains different classes and properties that represent the provenance information in different contexts, thereby usable for keeping track of the provenance information of linguistic corpora. The PROV ontology has been shown to be applicable to a diversity of fields to track provenance information in a general manner \cite{lebo2013prov}. 

Even though the RDF language is not optimized for this, the NLP Interchange Format (NIF)\footnote{\url{http://persistence.uni-leipzig.org/nlp2rdf/ontologies/nif-core}} makes it possible to represent texts and their elements in RDF, primarily by using strings. With the NIF ontology, a context, sentence, phrase, or word can be represented into RDF triples. It is also possible to have references between different levels, so a word can reference to another part of the RDF scheme that is the sentence from where it was derived. In a study by Menke et al. it was suggested to improve the reprensentation of the provenance information of the annotations in the NIF ontology by introducing a new ontology called MOND, which uses the PROV ontology to include provenance information in the NIF ontology \cite{menke2017origin}.


Moreover, many annotation guidelines exist and each is usually represented in a different structure, increasing the difficulty in creating a standard model to transform annotations into RDF. 
The Web Annotation Vocabulary \cite{sanderson2017oa}, for example, provides a possible solution with relations such as `motivatedBy' and `hasTarget', making it possible to structure annotations in an interoperable way.

Nanopublications are a format for small data publications based on RDF triples, consisting of three parts: the assertion, provenance, and publication information \cite{kuhn2013broadening}. The assertion contains the main content of a nanopublication, for example stating in a formal way the link between a gene and a disease. The provenance part states how this assertion came to be, by linking to the study that was performed or the paper from which the assertion was extracted. The publication information part, finally, records information about the nanopublication itself, such as by who and when it was created \cite{groth2010anatomy}. In a previous work, a Java library for nanopublications was created, which can also be used as a command line tool \cite{kuhn2015nanopub}. Nanopublications can be given \emph{trusty URIs} as identifiers, which include a hash value of the complete content, and thereby make nanopublications immutable and verifiable \cite{kuhn2014trusty}. Such nanopublications can then be reliably and redundantly published to the existing distributed server network \cite{kuhn2016decentralized}. Each of these servers contains all nanopublications, making it possible to retrieve the nanopublication from another server when one server is down. Nanopublications also allow for precise and reliable versioning, as well as the definition of incremental datasets by defining nanopublication indexes \cite{kuhn2017reliable}. Such indexes are represented as nanopublications themselves that contain links to other nanopublications and thereby defining sets of nanopublications.

\section{Methodology}
In this section we introduce our methodology to address the above-mentioned problems of interoperability and provenance of text corpora by applying the concept and techniques of nanopublications. To demonstrate and evaluate this methodology, we then present a case study. We introduce a number of questions on the combined annotations that can be answered in an integrated and automatic manner through SPARQL queries. Through the SPARQL queries we will show that data of multiple corpora can be retrieved using a single query, solving the interoperability issues.
\subsection{Nanopublication Model for Text Corpora} 

\begin{figure}[tb]
    \centering
    \includegraphics[width=0.75\textwidth]{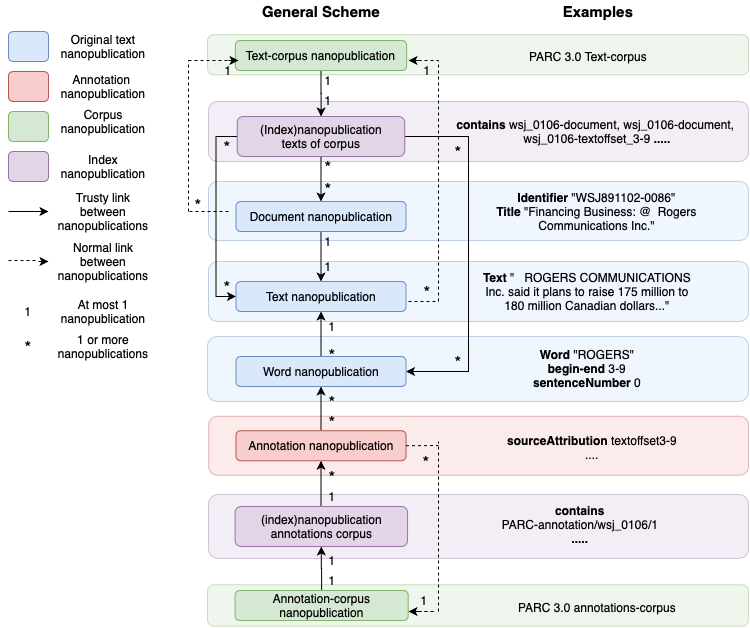}
    \caption{General nanopublication scheme with examples. }
    \label{fig:basicstructure}
\end{figure}

In Figure \ref{fig:basicstructure}, we present a general scheme for representing text corpora and their annotations as a network of interconnected nanopublications. 
This model ensures that new information from other corpora could easily be linked to existing nanopublications, through the addition of more nodes to the network. The nanopublications that form this general structure can be divided into four groups depending on what they represent: text, annotation, corpus, and index nanopublications. 



First, at the top of the figure the text-corpus nanopublication can be found. This nanopublication contains the information regarding the text corpus, which is the set of original texts. It is linked to an index nanopublication which points to all document, text, and word nanopublications which are part of the corpus. The document and text nanopublications can be part of multiple corpora and thus multiple index nanopublications. These document nanopublications contain all of the metadata on the actual texts, namely the provenance information, title, and identifier. Furthermore, the document nanopublications have a link to the nanopublication containing the actual text corresponding to the metadata. Both the document and the text nanopublication also contain a back-reference to their corpus nanopublication. 


Additionally, word nanopublications are created, as words are the typical targets of annotations. They contain the word string, sentence number, and the offset values with respect to the text, and possibly other word-level information (e.g. their part-of-speech value). Thereby, each word gets its own URI as identifier, which is constructed from the text URI and the word's offset values.
From the bottom, annotation corpora are pointing to these words. Similar to the text corpora, a corpus nanopublication of an annotation corpus contains the general corpus-level information, most importantly a link to an index nanopublication pointing to all annotation nanopublications.
Their concrete content depends on the type of annotation, but they all point to the words they are annotating via the URIs as described above. For convenience, they are also pointing back to the overall annotation corpus nanopublication.

Modeling these elements as small self-contained entities has also the advantage that differences in licenses can be defined and handled on a fine-grained level. Often, annotations have a different, sometimes more permissive, license than the texts they annotate. For this, we introduce the nanopublication type \texttt{ProtectedNanopub} that we use to mark nanopublications that cannot be openly published due to license reasons. We updated the nanopublication client and server code such that an error is raised if a user accidentially tries to publish such a protected nanopublication to the server network.

Lastly and importantly, our approach is open for anybody to publish further annotations or texts as nanopublications. These wouldn't be directly included in the datasets, as the index nanopublications don't point to them, but they can be found by querying the nanopublication network, and included if desired in a given situation. Moreover, a new version of a corpus can include such ``third-party'' contributions, thereby making them a proper part but also clearly attributing the third-party contributor, which is defined in the provenance and publication info parts. This also allows for corrections to be published by anybody who finds something that is wrong.

\subsection{Case Study Design}  

The case study workflow has 3 main components: (1) obtain and analyze original and annotated corpora; (2) convert these into nanopublications; and (3) assess the usefulness of the nanopublication representations. We evaluate on two event-related linguistic corpora, FactBank and PARC 3.0. For PARC, we process 34 annotation files that provide annotations on Wall Street Journal Corpus (WSJ) documents. For FactBank, we process 194 documents, which provide annotations on WSJ, New York Times (NYT), Associated Press Writer (APW) documents, and documents from several other sources. Thus, we use the intersection of 34 WSJ documents for which both corpora provide annotations for investigating the extent to which nanopublications indeed facilitate automated merging and interoperability.

The PARC corpus provides annotations about attribution relations, which are made up of three components a Source, a Content, and a Cue. The original annotations are made at the token-level and are presented in XML format. The FactBank corpus includes annotations about events and their factuality values. FactBank adopts event annotations from the TimeML annotations \cite{sauri2006timeml}, while it defines event factuality with a value that represents how certain or true an event is according to a source present in the text. FactBank has six primary factuality values present in its annotations. These annotations are contained in a set of 20 tables, each written in a separate text file.
We refer the reader to \cite{sauri2008FactBank} and \cite{pareti2015attribution} for detailed explanations of each annotation scheme, and we will elaborate below on their interoperability challenges.

For this case study, we came up with six general questions that somebody who would like to use the above annotations might want to ask. These questions are shown in Table \ref{case_study_questions}.
One of our goals is answering  questions that show the merging capabilities and efficiency of nanopublications. For example, the answer to  Q3, depending on the factuality values it outputs, might suggest how certain the content of the text is. An answer to Q6 might help an annotator clarify doubts during the annotation process, and Q5 demonstrates how FactBank events can be related to PARC attribution relations.
Some questions ask only for information from one corpus, while others are more complex by requiring information from both corpora.
We will use these general questions to make them automatically executable with our integrated nanopublication model and the SPARQL query language.

\begin{table}[tb]
\centering
\small
\begin{tabular}{r@{~~}p{14.8cm}}
 & Question \\
\hline
Q1 & Who talked about an event, and what is the factuality value of that event? \\
Q2 & Which events have multiple factuality value annotations? \\
Q3 & What are the different factuality values expressed per document, and which values appear the most? \\
Q4 & How often are certain annotation values used? (e.g. the count of different factuality values represented or the count of different sources)\\ 
Q5 & How many of the annotated FactBank events are labelled with a specific attribution component? \\
Q6 & Where in the corpus is a specific word (or lemma) assigned a specific attribution label (e.g. where is the verb `to surprise' annotated as Cue? \\
\end{tabular}
\caption{Case study questions.}\label{case_study_questions}
\end{table}

\section{Case Study} 
We applied the nanopublication-based approach to the case study in order to address the problem of interoperability and provenance. Different types of nanopublications were created to represent the chosen corpora. As introduced above, nanopublications consists of an assertion, provenance, and publication info part. 
In these nanopublications the publication info part is similar for all our nanopublications, containing the date of creation, license, ORCID identifiers of the creators, and sometimes further comments and website links.
All the text and word nanopublications, as well as the annotations of FactBank, are also marked as protected, due to their private licences. The remaining nanopublications are public and can therefore be published to the server network for public accessibility.
The dataset of all public nanopublications can be found online.\footnote{\url{https://github.com/ucds-vu/provcorp-model}}



We defined four corpus nanopublications for our case study: one for the text contained in the corpus and one for the annotations contained in the corpus, for each of the two existing corpora (PARC and FactBank). They contain basic information about the corpus and link to their content via an index nanopublication, as can be seen in this example (abbreviated here and below for readability):
\begin{lstlisting}[basicstyle=\scriptsize\ttfamily,frame=single,breaklines=true]
sub:assertion {
  corpus:parc-annotations a pvcp:AnnotationCorpus;
    dct:title "PARC Annotation corpus";
    rdfs:seeAlso <https://www.aclweb.org/anthology/L16-1619.pdf>;
    dcat:distribution indexnp: .
}
\end{lstlisting}
%
Via their index nanopublications, they point to 194 document nanopublications, which contain the document metadata. This is an example:
\begin{lstlisting}[basicstyle=\scriptsize\ttfamily,frame=single,breaklines=true]
sub:assertion {
  sub:document a foaf:Document;
    dct:title "Financing Business: @  Rogers Communications Inc.";
    dct:created "1989-11-02T00:00:00"^^xsd:dateTime;
    dct:creator <http://dbpedia.org/resource/The_Wall_Street_Journal>;
    pvcp:hasText textnp:text .
}
\end{lstlisting}
These document nanopublications link to the actual text using \texttt{pvcp:hasText} to point to the text nanopublication.
An example of the latter is shown here:
\begin{lstlisting}[basicstyle=\scriptsize\ttfamily,frame=single,breaklines=true]
sub:assertion {
  sub:text a nif:OffsetBasedString, dct:Text;
    rdf:value """   ROGERS COMMUNICATIONS Inc. said it plans to raise 175 million to 180 million Canadian dollars (US$148.9 million to $153.3 million) through a private placement of perpetual preferred shares. ... He declined to discuss other terms of the issue. """ .
}
\end{lstlisting}



The word nanopublications point back to the text nanopublication and provide word-level information including the word string, the offset of this string with respect to the text, and the sentence number. They can also contain lemma and part-of-speech information. As this word level depends on tokenization, there can be differences in how different annotation corpora define word boundaries. For this reason, our approach allows for the generation of new word nanopublications on demand. This creates an identifier for the word based on the URI of the text nanopublication and the offset of the word, thereby assuring that words introduced several times by different annotation corpora get the same identifier and are therefore immediately interoperable. In our case, we have 6,362 word nanopublications defined from PARC and 7,784 from FactBank. This is an example of the former:
\begin{lstlisting}[basicstyle=\scriptsize\ttfamily,frame=single,breaklines=true]
sub:assertion {
  textoffset:3-9 a nif:OffsetBasedString, nif:Word;
    nif:beginIndex "3"^^xsd:int;
    nif:endIndex "9"^^xsd:int;
    nif:anchorOf "ROGERS";
    nif:lemma "rogers";
    olia:POS "NNP";
    pvcp:hasSentenceNumber "0"^^xsd:int;
    pvcp:isPartOfText textnp:text .
}
\end{lstlisting}

The annotations can now link to these words. We have 136 annotation nanopublications for PARC, that associate words with content, cue and source annotations, as can be seen in the example below:
\begin{lstlisting}[basicstyle=\scriptsize\ttfamily,frame=single,breaklines=true]
sub:assertion {
  sub:annotation a oa:Annotation;
    dct:isPartOf corpus:parc-annotations;
    pvcpp:hasContentAnnotatedWord textoffset:101-106, textoffset:107-114, ...;
    pvcpp:hasCueAnnotatedWord textoffset:30-34;
    pvcpp:hasSourceAnnotatedWord textoffset:10-24, textoffset:25-29, textoffset:3-9 .
}
\end{lstlisting}
%
%
%
%
For FactBank, we have annotation nanopublications that declare events (7,784) and separate ones annotating their factuality values (10,948), with a similar structure as the PARC nanopublications above.\footnote{The complete set of nanopublications can be found here: \url{https://github.com/ucds-vu/provcorp-model}}

We have focused above only on the assertion part, but all these nanopublications also come with specific provenance and publication information, for example describing that the PARC annotations were created at a different time by a different person than the nanopublications themselves. 

\section{Results}
In this section we analyze the interoperability challenges we faced and addressed with our approach and assess the extent to which we can  automatically answer the questions we introduced above. 

\subsection{Analysis of Addressed Challenges} 

\begin{table}[tb]
\small
\centering
\begin{tabular}{rl@{~}|@{~}l@{~}|@{~}r@{~~}r}
 & Addressed Challenge & corpus & abs. & rel. \\
\hline
1. & Incompatible text offsets & PARC/FactBank & 194 & 100.0\% \\
2. & Metadata included in sentence number count & FactBank & 194 & 100.0\% \\
3. & Insufficient sentence splitting information & FactBank & 64 & 33.0\% \\
4. & Missing headline & FactBank & 33 & 17.0\% \\
5. & Inconsistent use of text tags & FactBank & 23 & 11.9\% \\
6. & Absence of text tags to structure the document & FactBank & 17 & 8.8\% \\
7. & Unknown journal / source & FactBank & 16 & 8.2\% \\
8. & No attribution relations for the document & PARC & 4 & 2.0\% \\
9. & Incompatible sentence splitting at semicolons & PARC/FactBank & 2 & 1.0\% \\
10. & Annotations on the headline & FactBank & 1 & 0.5\% \\
\end{tabular}
\caption{Overview of addressed interoperability challenges, including their absolute and relative frequencies in the 194 documents of our case study corpora.}
\label{tab:challenges}
\end{table}
With our case study as described above, we encountered a number of interoperability challenges. We managed to resolve almost all of them in our manual conversion work as described above. However, we had to exclude ten documents of FactBank due to the fact that sentence numbers were skipped when a sentence ended with a double quote (\texttt{"}), and we did not have enough information to correct this mistake. In Table \ref{tab:challenges}, we show an overview of the challenges we successfully addressed on the remaining 194 documents. This highlights the nature and extent of interoperability problems that come with the current way how such corpora are published but would be resolved if our approach was followed, i.e. if they would be published as interoperable nanopublications from the start.


The first issue on the list are text offsets. PARC and FactBank locate words in a text differently: PARC uses byte counts, whereas FactBank makes use of sentence and token numbers. Token numbers are tokenizer-specific, so FactBank's method is quite sensitive and difficult to replicate. FactBank's sentence numbers are moreover confusing as they also count the metadata fields, such as text headings, and not just the text content (which is challenge number 2). PARC, on the other hand, also has its quirks, such as including the six characters of the initial \texttt{<TEXT>} tag in the offset count.
In our nanopublications, we use text offsets that unambiguously link to the text string of the content, and thereby ensure precise and interoperable text references.



For FactBank's sentence and token number approach to refer to words in the text, it is crucial to know where a sentence ends and a new one starts. This sentence splitting, similar to tokenization mentioned above, is sensitive to the used algorithm. This is why text corpora like WSJ provide us with this sentence splitting information by saving each sentence on a separate line in the provided file formats. However, the documents from NYT and APW didn't have this information and so these sentences had to be checked manually for the correct splitting (challenge 3).
Moreover, sentences containing a semicolon in PARC are annotated as one sentence while they are considered two sentences in FactBank (challenge 9).

Challenges 5 and 6 point to the fact that text tags were used inconsistently. The main text is included in \texttt{<TEXT>} tags, but these were missing for 17 documents. Moreover, WSJ documents use  \texttt{<HL>} and \texttt{<DATELINE>} tags for the headline and date, respectively, while NYT 
and APW documents use \texttt{<HEADLINE>} and \texttt{<DATE\_TIME>}.
Some documents moreover, did not contain tags at all, and we had to add them manually.

Normally, headlines are not annotated, but there is an exception in document \path{APW19980213.1380}, which includes an annotation of a part of the headline (challenge 10). Upon encountering this situation, we extended our model to also allow for headline annotations.
%
Lastly, in some cases metadata was missing. Some documents did not have a headline (challenge 4) or did not include a journal or source (challenge 7).
In PARC, there were also four documents that did not contain any attribution relations (challenge 8).

In summary, all documents were affected by at least two challenges, and there seems to be a long tail of infrequent exceptions or small inconsistencies that one has to take into account in order to process such corpora correctly.

\subsection{Query Results} 
In order to assess the practical benefits of our approach we implemented the general questions presented in Table \ref{case_study_questions} as queries in the SPARQL language so we could use a triple store (Virtuoso in our case) to automatically answer them. The resulting SPARQL queries can be found online\footnote{\url{https://github.com/ucds-vu/provcorp-model/tree/master/queries}}.
%
%
%
They demonstrate that our nanopublication approach effectively merges the two corpora and allows for information to be extracted together in a practical and efficient way. Table \ref{query_stats} shows the average amount of time it took to run a query, as well as the number of rows it outputs.
We managed to represent all these questions in SPARQL, and thereby to make them automatically executable.

\begin{table}[tb]
\centering
\small
\begin{tabular}{r|l@{~}|@{~}r@{~~}r}
question & short query description & count & time\\
\hline
Q1 & list source and factuality value per event & 656 & 0.206s \\
Q2 & list events with more than one factuality value & 1,050 & 0.092s \\
Q3 & list factuality values per document & 194 & 0.058s \\
Q4 & count of FactBank source values & 425 & 0.029s \\
Q5 & count of FactBank events with PARC source attribution & 6 &  0.033s \\
Q6 & list all occurrences of ``surprise'' as cue & 1 & 0.048s \\
\end{tabular}
\caption{SPARQL query results and average execution times to answer the general questions Q1 to Q6}
\label{query_stats}
\end{table}

The listing below shows an example of a SPARQL query to answer one possibility for question Q4 (``How often are certain annotation values used?'') retrieving the count of words annotated as events from FactBank per PARC annotation attribution type:
\begin{lstlisting}[basicstyle=\scriptsize\ttfamily,frame=single,breaklines=true,morekeywords={prefix,select,count,distinct,as,where,values,group,by}]
prefix pvcpp: <https://w3id.org/provcorp/vocab/parc/>
prefix pvcpf: <https://w3id.org/provcorp/vocab/FactBank/>
prefix oa: <http://www.w3.org/ns/oa#>

select ?Attribution (count(distinct ?word) as ?Count) where {
  ?event pvcpf:hasEID ?eventid.
  ?event oa:hasTarget ?word.
  ?annotation ?Attribution ?word.
  values ?Attribution { pvcpp:hasContentAnnotatedWord pvcpp:hasCueAnnotatedWord pvcpp:hasSourceAnnotatedWord }.
} group by ?Attribution 
\end{lstlisting}
The concrete result for this query is:
\begin{lstlisting}[basicstyle=\scriptsize\ttfamily,frame=single,breaklines=true,morekeywords={Attribution,Count}]
Attribution                    Count
pvcpp:hasSourceAnnotatedWord       3
pvcpp:hasContentAnnotatedWord    284
pvcpp:hasCueAnnotatedWord        130
\end{lstlisting}
This thereby demonstrates how FactBank and PARC are conceptually merged and can be queried in an integrated fashion.

As another example, the query representing question Q1 (``Who talked about an event, and what is the factuality value of that event?'') gives us this result (only the first four rows are shown):
\begin{lstlisting}[basicstyle=\scriptsize\ttfamily,frame=single,breaklines=true,morekeywords={textID,eID,eventWord,factValue,relativeSource,sourcePhrase}]
textID                       eID  eventWord     factValue relativeSource   sourcePhrase
https://.../wsj_0026...#text e1   said          CT+       AUTHOR           The White House
https://.../wsj_0026...#text e11  requested     CT+       AUTHOR           Timex
https://.../wsj_0026...#text e123 beneficiaries CT+       officials_AUTHOR U.S. trade officials
https://.../wsj_0026...#text e123 beneficiaries Uu        AUTHOR           U.S. trade officials
...
\end{lstlisting}
This shows again how the annotations from both corpora can be queried simultaneously. The result is very informative for the file \path{wsj_0026}. The content of the columns, in order of appearance from left to right, is: the text identifier in the form of its nanopublication URI with the suffix \path{#text}, the event identifier of the event annotation, the word that is annotated as the event, the factuality value of the event, the source of the event (according to FactBank annotations), and the string in the document's text that represents the source (as annotated in PARC). As such, the query's result contains information from PARC (the source) as well as information from FactBank (the factuality value and event). Moreover, the output of the query allows us to see that an event can have two different FactBank sources who each give a different factuality value to the annotated event. This is seen in the last two rows where the event `beneficiaries' has two different factuality values, one which was given by the \texttt{AUTHOR} (the writer of the text document) and another given by the `officials' (a source talked about in the text document). The different conception of `source' in FactBank and PARC can also be noted here, which shows that although annotations can be merged, the annotation schema of each corpus should always be considered when analyzing the results. This is also useful to reveal the representation of a corpus's data more easily. For example, we can see that an event from FactBank always has an \texttt{AUTHOR} source. 

This demonstrates how we can use the power of query languages like SPARQL to access corpora in an integrated and fully automated way. Because the two corpora of our case study and the annotations are now formally linked, they are interoperable and can conceptually be seen as a single resource. Also, the results can be exported in a variety of formats, including CSV tables and JSON files, and be loaded and processed in other tools. It is then straightforward to write conversion tools to other formats, such as the CoNLL format \cite{buchholz-marsi-2006-conll}, a common practice data format used by computational linguists. This approach ensures that the data is interoperable from the start, and then format conversions are relatively easy as the range of nasty interoperability challenges is already taken care of.



\section{Discussion and Conclusion} 

Nanopublications have been mostly used with scientific data, such as data from the biomedical domain \cite{kuhn2013broadening}. In this work, we show that linguistic corpora can also benefit from the fine-grained and provenance-aware structure of nanopublications. The collection of nanopublications presented here contains information from event annotated corpora. It is demonstrated that when these data is modelled in a homologous way, each unique corpus can be automatically merged in order to produce valuable  data combinations that were not easily attainable prior to the merging. Specifically, our dataset can provide attribution annotation insight about events from FactBank.

Moreover, our nanopublications suggest the prospect of linguistic data fitting into the Linked Data ecosystem. During the dataset creation, we were able to reuse multiple existing Linked Data vocabularies. Linguistic data becoming more linked would also allow it to be connected more easily with already existing linguistic Linked Data sources from LLOD. Nevertheless, the data format of the original annotated corpora is extremely variable, which shows in the multiple textual challenges faced during production. While this project only focused on two corpora, there was inconsistency in annotation format not only between the corpora, but also within a corpus itself. Thus, the conversion of existing linguistic corpora into a Linked Data scheme can expect tedious pre-processing efforts. Even in our dataset, several documents needed to be excluded for the final version. Another aspect that makes it challenging to fit linguistic corpora in the Linked Data ecosystem relates to copyright availability. For example, this caused us to only be able to represent part of the PARC corpus.


In this paper, interoperability issues in linguistic data, specifically event annotated corpora, are discussed and nanopublications are proposed as a solution to resolve them. When one set of annotation guidelines is used to produce annotations on one corpus, that same corpus might already have other annotations that could be useful for the researcher. However, currently it is very difficult for a researcher to know or retrieve the version or provenance of these existing annotations due to poor documentation and variable annotation formats. Aspects of nanopublications that make them useful for annotated corpora are that they are able to be written in different formats, they can represent versioning, and they can track provenance. Also, they follow LLOD principles of using URIs for identification and RDF representation as output. Above all, they seem useful for merging linguistic annotation data. A successful model for translating existing event corpora into nanopublications is significant not only for computational linguistics, but also for the field of Linked Data.



\section*{Acknowledgements}

This research has been supported by the Academy Assistants Program from the Network Insitute from the VU Amsterdam. We are grateful to Silvia Pareti for making the PARC corpus available. 

\bibliographystyle{coling}
\bibliography{references.bib}

\end{document}